\documentclass[letterpaper, 10 pt, conference]{IEEEconf}
\IEEEoverridecommandlockouts
\usepackage[utf8]{inputenc} %
\usepackage{lineno, color}
\usepackage{url}            %
\usepackage{booktabs}       %
\usepackage{amsfonts}       %
\usepackage{nicefrac}       %
\usepackage{graphicx}
\usepackage{amsmath}
\usepackage{algorithm}
\usepackage{algpseudocode}
\usepackage{multirow}
\usepackage{xcolor, colortbl}
\usepackage{romannum}
\usepackage{cite}
\usepackage{hyperref}       %
\makeatletter
\let\MYcaption\@makecaption
\makeatother

\usepackage{times}

\usepackage[font=footnotesize]{subcaption}
\newcommand*{\ournn}{{CBF-induced neural function}}
\newcommand*{\ournnabbr}{{CBF-INF}}

\newcommand*{\ourcontroller}{{CBF-induced neural controller}}
\newcommand*{\ourcontrollerabbr}{{CBF-INC}}

\newcommand*{\ourmethod}{{neural CBF-induced neural controller-enhanced RRT}}
\newcommand*{\ourmethodabbr}{{CBF-INC-RRT}}

\title{\LARGE \bf
Efficient Motion Planning  for Manipulators with \\
Control Barrier Function-Induced Neural Controller}

\makeatletter
\let\@makecaption\MYcaption
\makeatother

\author{%
  Mingxin Yu$^{1}$, 
  Chenning Yu$^{2}$, 
  M-Mahdi Naddaf-Sh$^{3}$, 
  Devesh Upadhyay$^{3}$, 
  Sicun Gao$^{2}$, 
  and Chuchu Fan$^{1}$
\thanks{This work was supported by the National Science Foundation (NSF) CAREER Award \#CCF-2238030 and the MIT-Ford Alliance Program.}
\thanks{$^{1}$ Department of Aeronautics and Astronautics, Massachusetts Institute of Technology, USA 
    {\tt\small \{yumx35, chuchu\}@mit.edu}}
\thanks{$^{2}$ Department of Computer Science and Engineering, University of California, San Diego, USA 
    {\tt\small \{chy010, sicung\}@eng.ucsd.edu}}
\thanks{$^{3}$ The work was done when authors were at Ford Motor Company, USA 
    {\tt\small \{mahdinaddaf, deveshu\}@gmail.com}}
}

\begin{document}
\maketitle
\thispagestyle{empty}
\pagestyle{empty}

\begin{abstract}
    Sampling-based motion planning methods for manipulators in crowded environments often suffer from expensive collision checking and high sampling complexity, which make them difficult to use in real time. To address this issue, we propose a new generalizable control barrier function (CBF)-based steering controller to reduce the number of samples needed in a sampling-based motion planner RRT. Our method combines the strength of CBF for real-time collision-avoidance control and RRT for long-horizon motion planning, by using \ourcontroller\ (\ourcontrollerabbr) to generate control signals that steer the system towards sampled configurations by RRT.
\ourcontrollerabbr\ is learned as Neural Networks and has two variants handling different inputs, respectively: state (signed distance) input and point-cloud input from LiDAR.
In the latter case, we also study two different settings: fully and partially observed environmental information. 
Compared to manually crafted CBF which suffers from over-approximating robot geometry, \ourcontrollerabbr\ can balance safety and goal-reaching better without being over-conservative.
Given state-based input, our \ourmethod\ (\ourmethodabbr) can increase the success rate by $14\%$ while reducing the number of nodes explored by $30\%$, compared with vanilla RRT on hard test cases. Given LiDAR input where vanilla RRT is not directly applicable, we demonstrate that our \ourmethodabbr\ can improve the success rate by $10\%$, compared with planning with other steering controllers. Our project page with supplementary material is at \href{https://mit-realm.github.io/CBF-INC-RRT-website/}{https://mit-realm.github.io/CBF-INC-RRT-website/}.

\end{abstract}

\vspace{-4pt}
\section{Introduction}

Despite the wide adoption of robotic manipulators in many real-world applications, real-time planning of safe execution paths for manipulators in crowded environments can still be challenging due to high-dimensional complex dynamics. Sampling-based motion planning methods, such as Rapidly Exploring Random Trees (RRT)~\cite{lavalle2001rrt}, Probabilistic Roadmaps (PRM)~\cite{kavraki1996prm}, and their extensions~\cite{webb2013kinodynamic, kala2013rapidly, brunner2013hierarchical, wang2020neural, ichter2020learned}
have demonstrated their efficacy in generating collision-free paths in complex environments. However, 
the high sampling complexity of those methods is prominent for manipulators. Moreover, those methods require accurate state estimation before planning, which often assumes static environments and precise knowledge of the shapes and positions of obstacles. 

\begin{figure}[!t]
    \centering
    \includegraphics[width=0.7\linewidth]{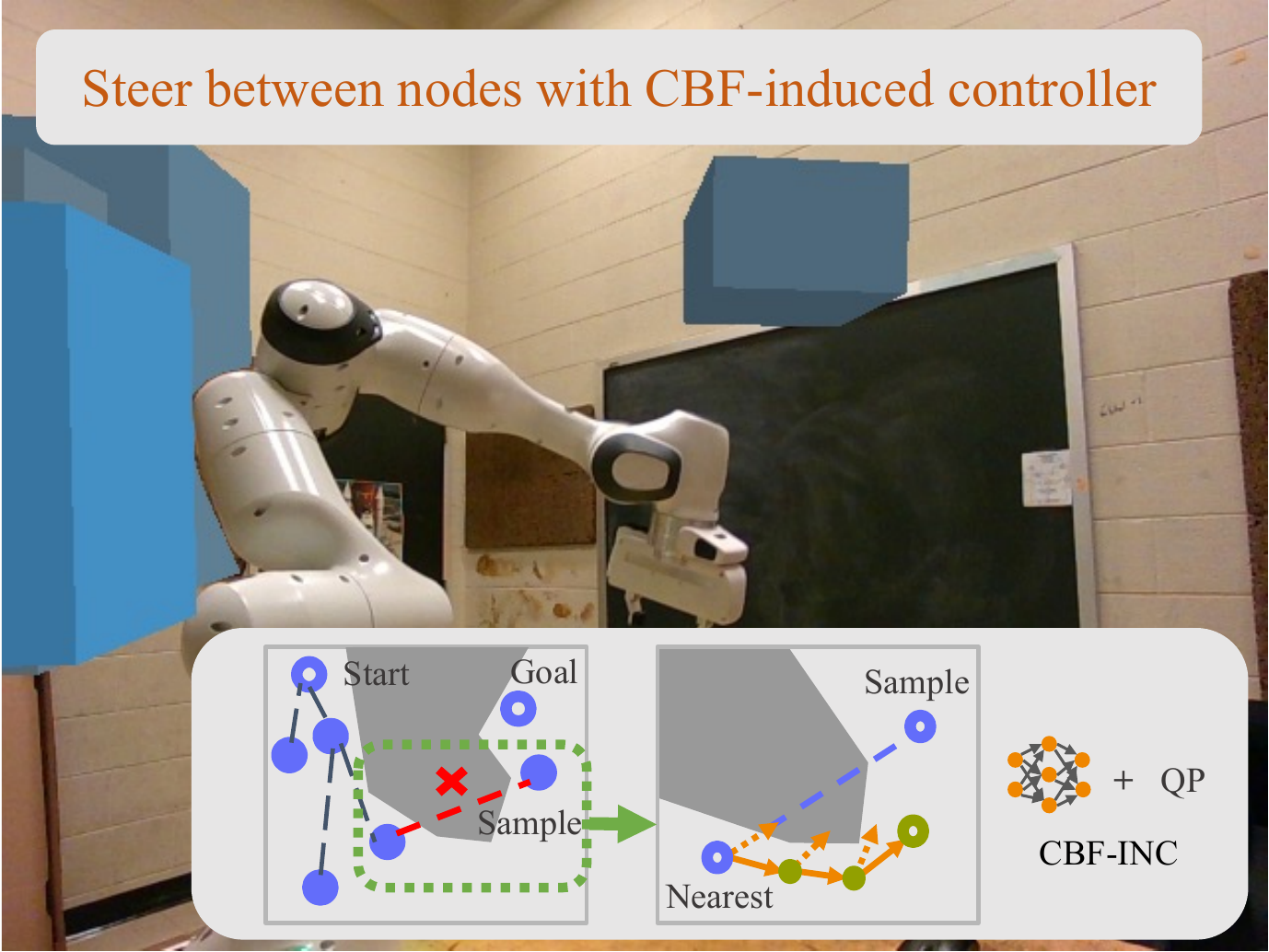}
    \caption{An example of a Franka Emika Panda robot planning motions using our proposed framework. We train a \ourcontroller\ and integrate it into the steer function in sampling-based motion planning. The controller is used for safe exploration and steers the edge to collision-free space without being over-conservative. } 
    \label{fig:teaser}
    \vspace{-18pt}
\end{figure}

The number of samples used in RRT can be reduced if expanding a node has a higher likelihood of success, which helps to reduce the node number for finding a solution. The expansion of a node will terminate when a collision is detected. On the contrary to recklessly heading towards the sampled state, using a safe steering function will substantially reduce the early termination of expansion.
In this paper, we are inspired by the use of control barrier functions (CBF)~\cite{AaronAmes2014, ames2016control} in multiple robotics applications for safe control to design such a steering function.
Prior works~\cite{manjunath2021safe, yang2019cbfrrt, ahmad2022adaptive, yang2023efficient} have shown significant advancements in combining the safe CBF controller with sampling-based motion planning algorithms for low-dimensional systems.
In the context of safe control, CBF has demonstrated success in rather complex robotic systems~\cite{wu2016safety, nguyen20163d}, including manipulators~\cite{singletary2022food, dai2023safe}. However, the CBFs used in the above works are typically manually designed as functions over the state of the environment, requiring both extensive experts' experience and accurate estimation from sensory data. When applying to robotic manipulator systems, the articulated nature of robots also poses an extensive computation burden for state estimation to handle the varying geometry of robots~\cite{landi2019ecc, singletary2022food}. 
For the simplicity of construction, geometric shapes of a robot are often over-approximated~\cite{cheol1994two, rimon1997obstacle, lin20166dof, landi2019ecc} and CBFs are often selected in simple forms like signed distance or ~\cite{landi2019ecc, singletary2022food} or in quadratic form~\cite{singletary2019manipulator, singletary2022kinematic}. The cost for simplicity is that the CBFs may be over-conservative and a feasible control signal is not guaranteed to be found.
To address these challenges, we are inspired by the recent advances in learning CBFs for the efficient synthesis of safe controllers, which has shown success in walking~\cite{castaneda2021gaussian}, flight~\cite{sun2021learning} and multi-agent setting~\cite{qin2021learning}, with some other attempts to extend these neural CBFs to observation-feedback systems~\cite{dawson2022hybrid, dean2020robust}. 

In this work, we build upon the motion planning framework RRT~\cite{lavalle2001rrt} and learn a \ourcontrollerabbr\ to steer the system towards newly sampled configuration. \ourcontrollerabbr\ has two variants handling different inputs: state (signed distance) and point cloud input from LiDAR. Given state input, our framework \ourmethodabbr\ increases the success rate by $14\%$ and reduces the number of explored nodes by $30\%$ on the most challenging test cases, compared with vanilla RRT and other neural-controller-enhanced RRT. \ourmethodabbr\ also doubles the success rate and halves the explored nodes, compared with the hand-crafted CBF-enhanced RRT method by avoiding over-conservativeness. 
With point cloud input setting, where many methods (like vanilla RRT and hand-crafted CBF) are not directly applicable, \ourmethodabbr\ still improve the success rate by $10\%$ on challenging cases, compared with planning with other steering controllers.

\noindent\textbf{Contributions.} Our contributions are summarized below:
(\romannum{1}) We present a \ourcontrollerabbr\ specialized for robotic manipulators. Our neural networks tackle high-dimensional observations and complex geometric link shapes. To the best of our knowledge, this is the first CBF-style controller taking raw sensor input in high-dimensional manipulators.
(\romannum{2}) We present a framework - \ourmethodabbr\ that incorporates the learned neural CBF into the motion planning algorithm. Such a framework preserves the completeness of traditional motion planning methods while benefiting from the safe exploration of \ourcontrollerabbr.
(\romannum{3}) Through extensive experiments on 4D and 7D manipulators, we demonstrate that planning algorithms using \ourcontrollerabbr\ significantly outperform baselines, in terms of success rate and exploration efficiency. We also show \ourcontrollerabbr\ generalizes to dynamic environments and evaluate \ourmethodabbr\ on hardware.

\vspace{-1pt}
\section{Related Works}\label{related}
    \noindent\textbf{Local Safety for robotic arms.} 
Safe deployment in the real world is crucial for general-purpose robotic arms. Various non-learning techniques have been proposed to tackle the collision avoidance problems on manipulators, ranging from potential field methods~\cite{Khatib1985realtime, Santis2007self, flacco2012depth}, reachability analysis~\cite{holmes2020reachable}, to CBF~\cite{landi2019ecc, dai2023safe}, a trusted tool for ensuring safety in control systems~\cite{wieland2007constructive, AaronAmes2014}. 
These methods, including CBF are usually hand-crafted via signed distance~\cite{landi2019ecc, haviland2020neo, singletary2022food}, quadratic form~\cite{singletary2019manipulator, singletary2022kinematic} or minimum uniform scaling factor~\cite{dai2023safe}. 
While effective for certain environments, they require substantial design efforts and perfect knowledge of the environment.
For simplicity, CBFs designed for robotic arms with multiple degrees of freedom (DoF) often use a set of simple convex shapes to over-approximate the rigid body links~\cite{cheol1994two, rimon1997obstacle, lin20166dof, landi2019ecc}, leading to over-conservative policies. Extending these approximations from one shape to another is also not straightforward.

In contrast, data-driven methods, exploiting the power of neural networks, have shown promise in addressing complex geometric shapes of manipulators~\cite{pham2018optlayer, nadia2023sdf}, even when only high-dimensional observations are available~\cite{liu2022safe}. 
Recently, learning-based CBFs have demonstrated potential in resolving these challenges on drones in multi-agent setting~\cite{qin2021learning} and visual navigation systems~\cite{dawson2022hybrid}. 
However, in robotic manipulator systems, the adoption of neural networks in CBFs is restricted to parameter search guidance~\cite{mcilvanna2022reinforcement} due to the convoluted collision-free configuration space. 

\noindent\textbf{Inductive bias in sampling-based motion planning.}
One stream of methods~\cite{ichter2018learning, jurgenson2019harnessing, strudel2021obstaclerep, zhang2022learning} has been proposed to improve sampling-based planning methods by incorporating inductive bias. The methods fall into two categories. One set of methods seeks to find heuristic functions to prioritize the samples to explore, including Fast Marching Trees~\cite{janson2015fast}, sampling-based A*~\cite{persson2014sampling}, and recent GNN-related works~\cite{yu2021reducing, zhang2022learning}. 
Some works also consider improving the sampling strategy~\cite{ichter2018learning, zhang2018learning}.
However, these methods still suffer from finding a control policy for the planned reference trajectory, especially for complex dynamics. 
The other class of methods conceives motion planning problems as sequential decision-making problems and relies on neural policies~\cite{pfeiffer2017perception, strudel2021obstaclerep, zhang2022learning} such as imitation learning~\cite{qureshi2019motion} or reinforcement learning~\cite{jurgenson2019harnessing} in an end-to-end manner, aiming to find a collision-free trajectory directly with a neural network. Some further consider adding explicit safety constraints during training to accelerate~\cite{pham2018optlayer, wang2022ensuring}.
Though these methods have shown impressive results, they sacrifice the completeness guarantee of many motion planning algorithms.

The most related works to this paper are~\cite{yang2019cbfrrt, ahmad2022adaptive, manjunath2021safe, yang2023efficient}, which explore incorporating CBF into sampling-based motion planning. ~\cite{yang2019cbfrrt, ahmad2022adaptive, yang2023efficient} proposed improvements on sampling methods and ~\cite{ahmad2022adaptive, yang2023efficient} focused on optimal planning. 
Our method only modifies the steer function in motion planning, and is therefore able to work directly with many sampling-based planners. While~\cite{yang2019cbfrrt, ahmad2022adaptive, yang2023efficient} work in simple 2D environments with a hand-crafted CBF and~\cite{manjunath2021safe} work on 2D car dynamics, we propose a \ourcontroller\ component and conduct experiments with high-DoF robotic manipulators in a highly cluttered environment.
Another closely related work is~\cite{nadia2023sdf}, which synthesizes a control policy for high-DoF manipulators via a neural signed distance function and quadratic programming (QP). The training of~\ournn\ explicitly penalizes the infeasibility of QP, improving the goal-reaching performance of controllers.

\vspace{-4pt}
\section{Preliminaries}
    We consider a robot with control-affine dynamics $\dot q = f(q) + g(q) u$, where $q\in\mathcal C\subseteq \mathbb R^n$ is the robot configuration and $u\in\mathcal U\subset \mathbb R^m$ is the control input, in a cluttered environment $\mathcal E$. We assume both configuration space $\mathcal C$ and action space $\mathcal U$ to be bounded. The robot perceives the environment $\mathcal E$ through an observation model $o = o(q, \mathcal E)\in\mathcal O\subset \mathbb R^k$. In this work, we consider two different observation models, namely, a signed-distance observation model and a LiDAR-based observation model, whose details are elaborated in~\ref{subsec:oCBF}. 
The state-observation space $\mathcal X:=\mathcal C\times\mathcal O$ can be partitioned into three subspaces: an unsafe set $\mathcal X_u$ where the robot collides with or penetrates itself or environmental obstacles, a safe set $\mathcal X_s=\{x| \|x-x_u\|\ge r_{\text{thres}}, \forall x_u\in \mathcal X_u\}$ where the robot is at least $r_{\text{thres}}$ away from the unsafe set, and a boundary set $\mathcal X_b=\mathcal X\setminus(\mathcal X_u\bigcup \mathcal X_s)$.

Given a start configuration $q_0$ and goal configuration $q_g$, satisfying $(q_0, o(q_0, \mathcal E)),\ (q_g, o(q_g, \mathcal E)) \notin \mathcal X_u$, we seek to find a feasible control sequence $\mathbf{u}:[0,T]\rightarrow \mathcal{U}$ that steers the system from $q_0$ to $q(T)\in\mathcal X_{\text{goal}}$, while ensuring $(q(t), o(q(t),\mathcal E))\notin \mathcal X_u, \forall t\in[0,T]$. The goal region $\mathcal C_{\text{goal}}$ is defined as $\{(q,o(q,\mathcal E))| \|q-q_g\|\le r_{\text{goal}}\}$ for some pre-defined radius $r_{\text{goal}}\ge0$.
In this work, we build upon motion planning algorithm RRT~\cite{lavalle2001rrt} and substitute the steer function with a neural-network-based controller for control input.

\noindent\textbf{Rapid-exploring random trees (RRT).}
RRT~\cite{lavalle2001rrt} tackles the motion planning problem by starting with sampling a set of configurations in the free space $\mathcal X_f=\mathcal X_s\bigcup\mathcal X_b$. The algorithm then attempts to build or expand a tree to connect these nodes with the start configuration $q_0$ and the goal configuration $q_g$. This two-step process is repeated until a collision-free path connecting the two configurations is found or until termination. Each edge on the tree has to be collision-free, thus requiring collision checking at the edge construction stage.

\noindent\textbf{Control barrier function.}
CBF ensures safety in control systems by enforcing the states of the systems to stay in the safe set. Extended CBFs in state-observation space $\mathcal X$ are scalar functions $h:\mathbb R^n\mapsto \mathbb R$ such that for some $\alpha_h>0$:
\begin{equation}
    \label{eq:cbfcondition}
    \begin{aligned}
    & \forall (q,o) \in \mathcal{X}_s, h(q,o) \leq -\gamma\\
    & \forall (q,o) \in \mathcal{X}_u, h(q,o)>\gamma \\
    & \forall (q,o) \in\mathcal X, \ \inf_{u\in\mathcal U} (L_fh(q,o) + L_gh(q,o)u)+\alpha_hh(q,o) \leq -\epsilon
    \end{aligned}
\end{equation}
where $L_fh$ and $L_gh$ denote the Lie derivatives, which capture the rate of change of $h$ along the system trajectories induced by $f$ and $g$, respectively. Since $o$ is also a function of $q$, the calculation of Lie derivatives requires the calculation of $\frac{\partial h}{\partial q}\dot q$ and $\frac{\partial h}{\partial o}\dot o$.
And $\epsilon, \gamma>0$ are small margins to encourage the strict inequality satisfaction of CBF conditions. 
It is proved in~\cite{AaronAmes2014} that if the initial state $(q(0),o(0))\in \mathcal{X}_s, h(q,o) \leq 0$ and a Lipschitz continuous policy $\pi:\mathcal X\mapsto\mathcal U$ selects actions from the set $\mathcal K_\text{CBF}=\{u \mid L_fh(q,o) + L_gh(q,o)u+\alpha_hh(q,o) \leq 0\}$, then the trajectory $q(\cdot)$ does not leave the safe set $\mathcal X_s$.

Here we can see a connection between motion planning and control barrier function. As long as the controller ensures the forward-invariant of the safety set, then the agent never encounters collisions. Using such a CBF controller ensures the collision-free constraint for motion planning.

\vspace{-2pt}
\section{Methods}
\vspace{-1pt}
    
Motivated by the connection between the motion planner's collision-free constraint and CBF's safety guarantees, we present a two-stage approach using a \ourcontroller\ that allows a robot to avoid obstacles and a motion planner that guides the robot to jump out of stuck regions and toward its goal. We first train a neural network \ournn\ (\ournnabbr) for the robot. \ournnabbr\ is trained to satisfy all the constraints in~\eqref{eq:cbfcondition}. Next, we synthesize a controller \ourcontrollerabbr\ using the trained \ournnabbr\ and incorporate it into the motion planning framework.

\subsection{Learning Framework for \ournnabbr}\label{method:trainCBF}
\noindent\textbf{Training procedures.}
We utilize an offline strategy to train \ournnabbr\ $h_\theta(q,o)$, where the training data is pre-collected. Our training dataset is a combination of two different parts, both are collected in various training environments : 
(\romannum{1}) We gather rollout trajectories using classical controllers (e.g. LQR controller). These trajectories are generated with random initial and goal states.
(\romannum{2}) To ensure coverage of less-explored spaces within the rollout, we uniformly sample the robot's pose in the configuration space. 
The presence of observation allows that the training environments not necessarily be the same as test ones and that \ournnabbr\ can easily generalize to new environments. \ournnabbr\ is trained to minimize an empirical loss function $\mathcal L$ like~\cite{dawson2022hybrid}:
\begin{equation}
\label{eq:loss}
\vspace{-2pt}
    \begin{aligned}
        &\mathcal{L}  =\dfrac{\alpha_1}{N_\text{safe}}\sum_{(q,o) \in \mathcal{X}_{s}} [\gamma+h\left(q, o\right)]_+\\
        &+\dfrac{\alpha_2}{N_\text{unsafe}}\sum_{(q,o) \in \mathcal{X}_{u}} [\gamma-h\left(q, o\right)]_+
        +\dfrac{\alpha_3}{N}\cdot \\&
        \sum_{(q,o) \in \mathcal{X}}
        [\epsilon + L_f h(q,o) + \inf_{u\in\mathcal U} (L_g h(q,o)\cdot u)
            +\alpha_h h]_+
    \end{aligned}
    \vspace{-4pt}
\end{equation}
where $\alpha_1, \alpha_2, \alpha_3$ are positive tuning parameters, $[\cdot]_+$ is $\max(0,\cdot)$, $N_\text{safe}$, $N_\text{unsafe}$ and $N$ are the number of points in the training samples in $\mathcal X_s$, $\mathcal X_u$ and $\mathcal X$, respectively.
As the term $\inf_{u\in\mathcal U} (L_g h(q,o)\cdot u)$ is linearly dependent on $u$, the minimum value can be easily found within a bounded action space $\mathcal U$ via linear programming (LP). The existence of a feasible control signal in the last condition in~\eqref{eq:cbfcondition} can be demonstrated when the third loss term comes to $0$. 
Note that \ournnabbr\ is trained as a Neural Network from finite samples and therefore not a valid CBF before being verified to satisfy the CBF constraints over the entire space. The latter is a theoretically hard problem~\cite{huang2020survey}. However, the learned \ournnabbr\ can be used as a steering function and provides significant empirical improvements over vanilla RRT (shown in Sec.~\ref{result}). We will also show in Fig~\ref{fig:learned-cbf} that the empirical satisfaction rates of the CBF constraints over finite samples are close to $100\%$.

\noindent\textbf{Computing Lie-derivatives.}
Directly computing the third condition in Eq.~\eqref{eq:cbfcondition} requires the calculation of $\frac{\partial h}{\partial q}\dot q$ and $\frac{\partial h}{\partial o}\dot o$. 
We first assume the obstacle velocities are much smaller than the links of manipulators in the dynamic scenarios, so we can disregard the change of $o$ in between two computational steps induced by the change of environment $\mathcal E$ when computing Lie-derivatives.
Furthermore, we replace the exact calculation of $\frac{\partial h}{\partial q}$ with numerical differentiation, i.e., $[\frac{\partial h}{\partial q}]_i=\frac{h(q+e_i\cdot\epsilon,o)-h(q,o)}{\epsilon}$, where $e_i$ is a one-hot vector with $e_i[i]=1$. This approach bypasses the explicit expression of forward kinematics of manipulators with many degrees of freedom and only demands a "black-box" access to the numerical values of the kinematics and the Jacobian~\cite{singletary2019manipulator}. 

\subsection{Specializing Functions for Manipulators: State-based and LiDAR-based \ournnabbr}\label{subsec:oCBF}

In contrast to hand-crafted CBFs utilized on multi-DoF robotic manipulators in ~\cite{yang2019cbfrrt, landi2019ecc, singletary2022food}, we aim to learn a neural function that encodes the safety constraint of avoiding both self-collision and collision with the exterior environment. Based on different observation models, we propose two types of \ournnabbr, both sharing the same training procedure.

\begin{figure}[t]
    \centering
    \begin{subfigure}[t]{0.29\linewidth}
        \centering
        \includegraphics[width=\textwidth]{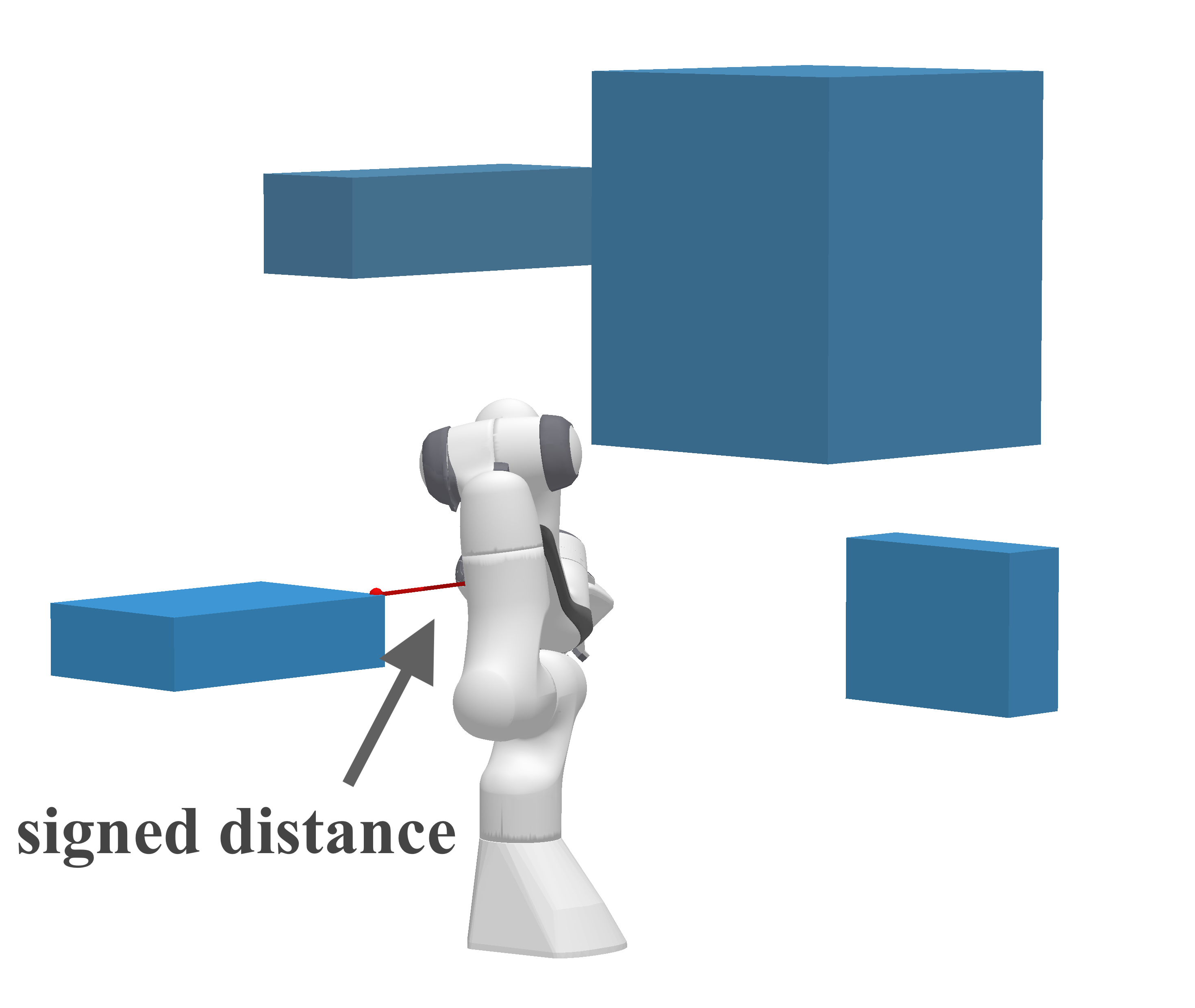}
    \end{subfigure}%
    ~ 
    \begin{subfigure}[t]{0.29\linewidth}
        \centering
        \includegraphics[width=\textwidth]{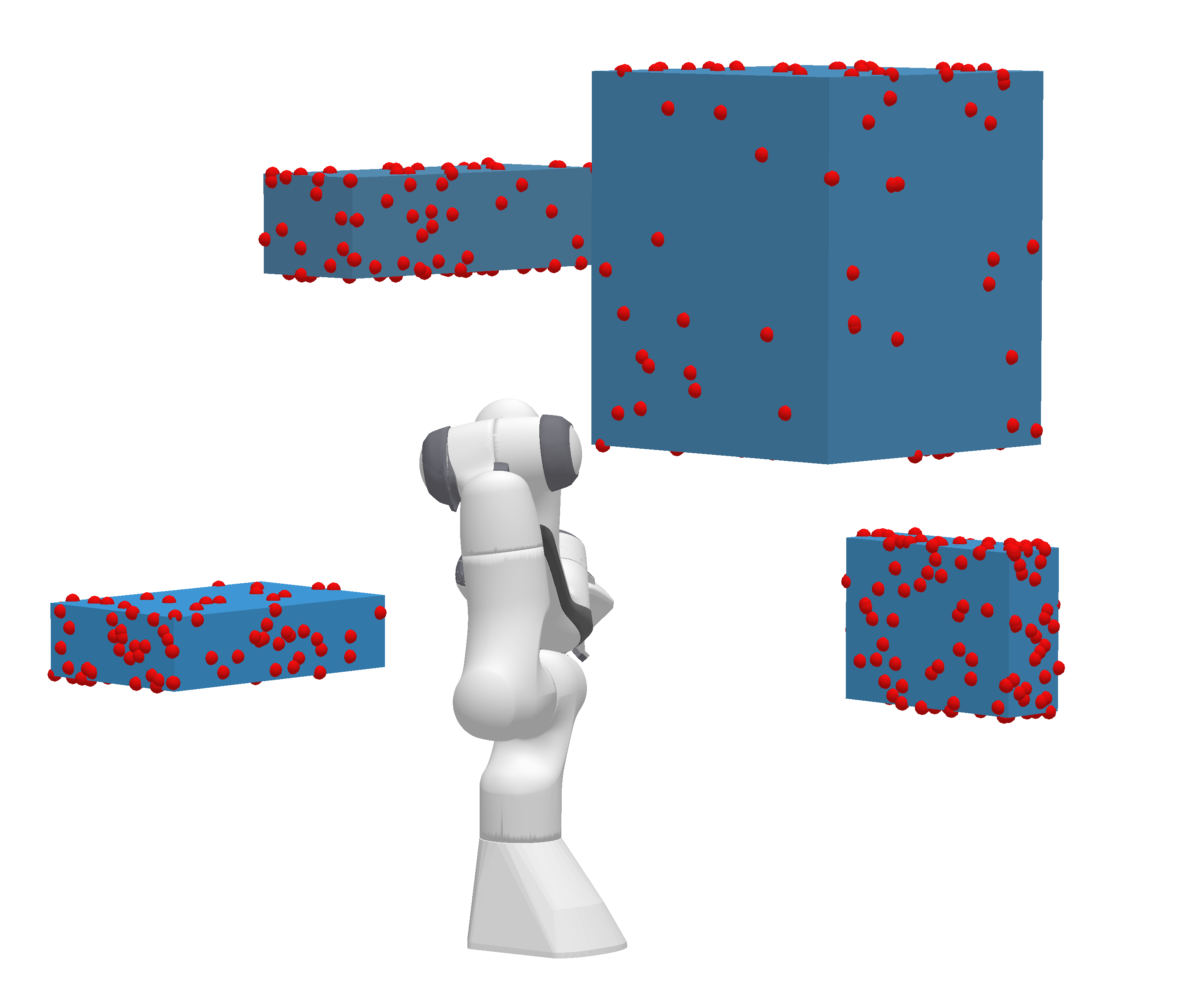}
    \end{subfigure}
    ~ 
    \begin{subfigure}[t]{0.29\linewidth}
        \centering
        \includegraphics[width=\textwidth]{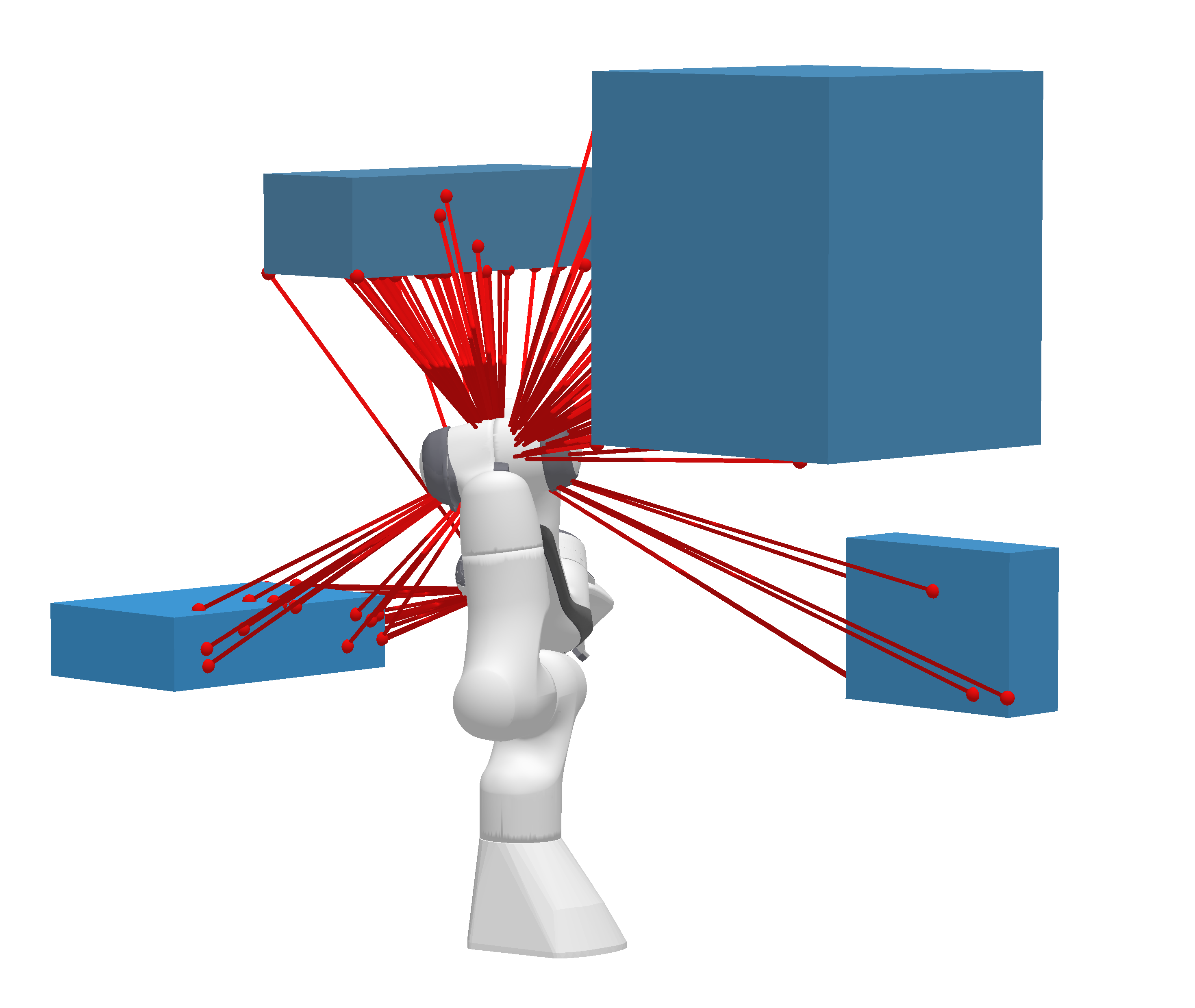}
    \end{subfigure}
    \caption{Illustration of observations.
    Left: s\ournnabbr\ takes the signed distance to the nearest obstacle as observation.
    Middle: For {\em fully-observable} environments, o\ournnabbr\ uses a point cloud sampled uniformly on the obstacles as observation.
    Right: For {\em partially-observable} environments, o\ournnabbr\ observes the point cloud from a mounted LiDAR sensor.}
    \label{fig:observation}
    \vspace{-12pt}
\end{figure}

\begin{figure}[!t]
    \centering
    \includegraphics[width=0.9\linewidth]{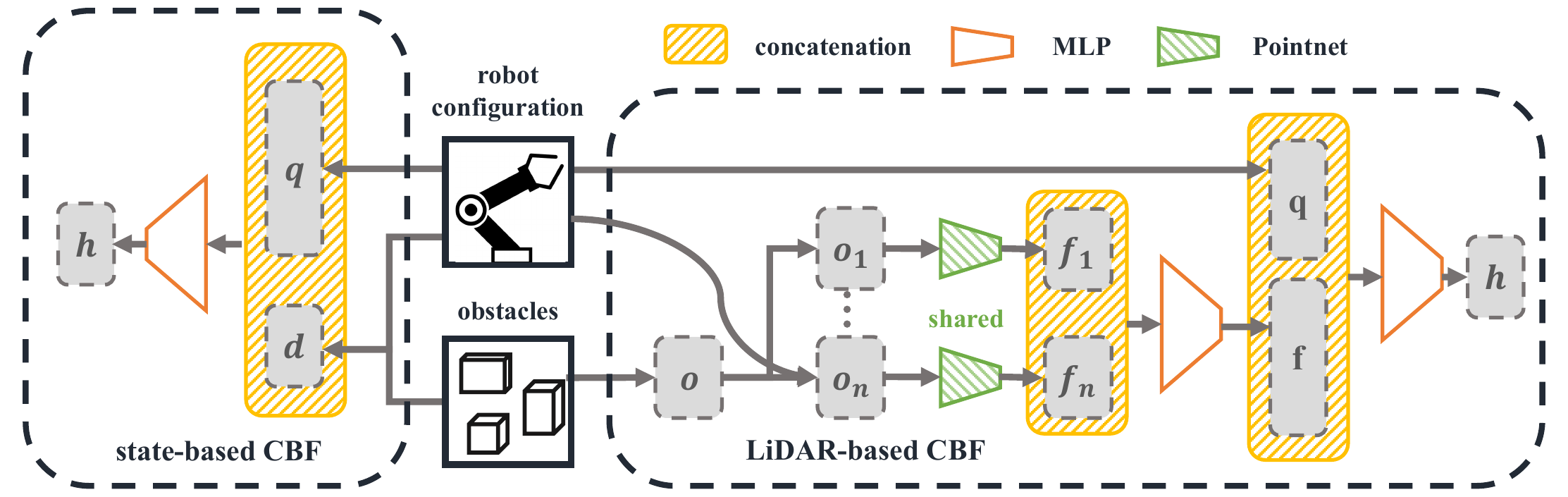}
    \caption{The overall neural network architecture. Left: The architecture of the s\ournnabbr. Right: The architecture of the o\ournnabbr.}
    \label{fig:nn_architecture}
    \vspace{-12pt}
\end{figure}

\noindent\textbf{State-based \ournnabbr\,(s\ournnabbr).} 
The observation $o$ in s\ournnabbr\ is 
the minimum signed distance, $d$, between the obstacles and any robot links. 
In this setting, it is easy for the \ournnabbr\ to determine, from the sign of $d$, whether the robot is in a collision-free state with respect to the environment. The only remaining challenge is to learn the self-collision pattern, which solely depends on the configuration $q$. 
To address this, we design 
the neural network to take the concatenated vector $q$ and $d$ as input, and generate a scalar as output, as in Fig.~\ref{fig:nn_architecture}. 

\noindent\textbf{LiDAR-based \ournnabbr\,(o\ournnabbr).}
s\ournnabbr\ relies heavily on accurate distance information from the environment, which is not available or costly to acquire in a dynamic or partially observable environment. A more flexible implementation is to define \ournnabbr\ as functions of partial observations, e.g., LiDAR.
In the LiDAR-based setting, the observation $o$ is composed of $N$ raw points sampled on the surface of obstacles paired with their respective normal vectors, similar to~\cite{strudel2021obstaclerep}. This observation is represented by a finite set $o=\{(p_i, n_i)\}_{i\in[1,\cdots,N]}\in\mathbb R ^{N\times 6}$, with the 3 dimensional points $p_i$ and 3 dimensional normal vector $n_i$ in the world frame. The point cloud can be retrieved using the LiDAR sensor or a depth camera mounted on the robot.

For each link of the manipulator, we transform the point cloud into its local frame, concatenate each point with a one-hot vector of the link index, and then feed all the transformed point clouds into a PointNet~\cite{qi2017pointnet}, which encodes the point clouds while ensuring permutation invariance on the order of points. The feature vectors from the PointNet are further fed into an MLP, concatenated with the configuration $q$. The whole network architecture is shown in Fig.~\ref{fig:nn_architecture}.

\subsection{Integrating \ournnabbr\ with Motion Planning}\label{method:pluginCBF}
Sampling-based motion planning algorithms are widely adopted to efficiently search high-dimensional spaces via building a space-filling tree. Despite their guaranteed completeness, these algorithms explore the configuration space via random shooting, and fine-grained collision-checking is required for each edge. The edge will be discarded if any part of the checking fails, even though a slight detour may save the edge, which wastes the computation in such a failed exploration.
Our framework, however, encourages the planner to explore the configuration space more wisely by using safe controllers, i.e., \ourcontrollerabbr. By incorporating the controller into the motion planning algorithm, the likelihood of successfully expanding a node is increased, which reduces the exploration cost given the controller's reactivity to obstacles.

\noindent\textbf{Synthesize controller.} We construct our controller \ourcontrollerabbr\ from CBF theory~\cite{AaronAmes2014} that modifies any given reference controller $u_{\text{nominal}}$ by solving the quadratic programming (QP) problem:
\begin{equation}
    \begin{aligned}
        &u(q,o,q_g)=\arg \min_{u\in\mathcal U} \|u-u_{\text{nominal}}(q,o,q_g)\|^2,\\ 
        \text{s.t.}&\ L_fh(q,o)+L_gh(q,o)\cdot u+\alpha h\le0.
    \end{aligned}
    \label{eq:qp}
\end{equation}
Here, the controller seeks to minimize the squared difference with the nominal control input within the action space $\mathcal U$, while satisfying the safety constraint.

\noindent\textbf{Safe-steering RRT.}
In this study, we focus on Rapidly-exploring Random Trees (RRT)~\cite{lavalle2001rrt} as a representative sampling-based motion planner. In RRT, the steer function generates a prospective edge, which extends the current search tree toward the direction of a randomly selected point. The steer function needs to conduct collision checking in the process.
Different from~\cite{yang2019cbfrrt}, which substitutes explicit checking for the nearest neighbor and changes the original framework, we propose \ourmethodabbr, to use \ourcontrollerabbr\ as the steer function. This is a general approach that can be applied to any sampling-based motion planners using a steer function. 
Within our steer function, the robot rolls out a trajectory using \ourcontrollerabbr. The $q_g$ in~\eqref{eq:qp} is set to be the newly sampled point. The generated trajectory and control sequence then serve as the edge added to the search tree. Details of the complete algorithm are provided in the Appendix.

There may be concerns about the completeness of this modified planning paradigm. However, we can still ensure completeness by opting to use \ournnabbr\ to discard unsafe LQR actions instead of modifying them after a certain number of exploration steps, as suggested in~\cite{yang2023efficient}. This optional variation allows us to maintain the crucial aspect of completeness while improving safety and efficiency.

\section{Experimental Results}\label{result}
    \noindent\textbf{Experiment setup.}
We evaluate our methods on a 4-DoF Dobot Magician in simulation and a 7-DoF Franka Panda both in simulation and the real world. Similar to~\cite{singletary2022food}, we consider direct control over the joint velocities, i.e., $\dot q=u$. In experiments, obstacles within the environment are depicted as cuboids. \ournnabbr\ is trained in environments with 4 fixed-size obstacles with random poses. We test the method on more challenging environments with 8 obstacles of random sizes and poses unless specified otherwise. The nominal policy $u_\text{nominal}$ for the QP controller is selected as LQR.
The simulations of continuous-time robot dynamics and control frequency occur at $120\,\text{Hz}$ and $30\,\text{Hz}$, respectively.

\noindent\textbf{Baselines.} Apart from vanilla RRT (RRT)~\cite{lavalle2001rrt}, we also compare against RRT variants with the following steer controllers in both state-based and LiDAR-based settings:
\begin{itemize}
    \item \textbf{Reinforcement Learning (sRL\&oRL)\cite{lillicrap2015continuous}:} We design the reward function to encourage goal-reaching and penalize collision, then train the controller with DDPG.
    \item \textbf{Imitation Learning (sIL\&oIL)\cite{torabi2018behavioral}:} Use behavior cloning to mimic the planned trajectories generated from an expert motion planner BIT*~\cite{gammell2014bit}.
\end{itemize}
Some baseline methods require complete information of the environment, thus only available in the state-based setting:
\begin{itemize}
    \item \textbf{Hand-crafted CBF (hCBF)~\cite{dai2023safe}:} The construction of this CBF adopts the minimum uniform scaling factor. Only available for 7-DoF Panda robot.
    \item \textbf{Safe RL method OptLayer (sOpt)~\cite{pham2018optlayer}:} Add additional optimization layer to force the controller satisfy~\ref{eq:cbfcondition}, where $h$ is a signed distance function instead of CBF.
\end{itemize}
The baseline methods with steer controllers are abbreviated with a suffix '-steer'. We also conduct ablation studies evaluating the controllers only.
All neural controllers are designed and trained using the same environment observations and neural network architectures.
The methods are evaluated on randomly generated 1000 easy and 1000 hard testing cases, based on the time required for BIT*~\cite{gammell2014bit} to find a solution. All the experiments are conducted with a predefined node limit: for the 4-DoF robot, exploration is restricted to a maximum of 200 nodes, while the limit for the 7-DoF robot is 500 nodes.

\subsection{Motion Planning in Simulation}\label{sec:planning_result}
\noindent\textbf{Evaluation metrics.} For motion planning problems in the section, we consider the following metrics: 
(1) Success rate (SR): A problem is successfully solved only if a collision-free path is found within the node limit. 
(2) Explored nodes: the attempts of adding a node to the search tree, regardless of success or not. 
(3) Total time consumption: One common concern about learning-based methods is their running speed due to the frequent calling of a large neural network model at inference time. We separate this metric into two categories: 
(3.1) online time, which is directly related to control frequency and observation update frequency and must be performed online during execution. This includes neural network inference time, QP solving, and perceiving observations; and 
(3.2) planning time, proportional to the total timesteps when expanding the search tree, is the remaining time consumption other than online time. This includes planning, running the simulations, and performing collision checking.
Because online time highly depends on selected parameters, we only report planning time in the main text. Results for online time can be found in the supplementary.

\noindent\textbf{Performance of state-based methods.} 
Shown in Table~\ref{tab:state_whole}, we see significant improvement in success rate and exploration efficiency (explored nodes) using s\ourmethodabbr\ in the state-based setting. Remarkably, performance improvement is much more pronounced on challenging hard testing problems. Regarding planning time, s\ourmethodabbr\ performs comparably with vanilla RRT and takes considerably less time compared to sRL-steer, sIL-steer methods, and even safe method sOptLayer-steer. 
It's worth discussing why hCBF-steer performs much worse than \ourmethodabbr\ and even RRT. First, hCBF is more conservative because it over-approximates the geometry shapes of the robot. This limits its performance, especially in cluttered environments. Second, the QP controller in hCBF-steer sometimes cannot find a feasible solution. Although we've attempted to relax the optimization problem with a constraint violation penalty term, this compromises the safety guarantee of hCBF-steer.

\begin{table}[!t]
    \centering
    \caption{Experiments under state-based setting. Performance on average success rate (SR), number of explored nodes on 1000 test cases, and summed planning time on 100 testing cases, averaged over 3 random seeds.}
    \label{tab:state_whole}
    {\setlength{\tabcolsep}{3.5pt}
    \begin{subtable}{0.5\textwidth}
  \caption{Results on 4-DoF Magician robot.}
  \label{tab:state-magician}
  \vspace{-1pt}
  \centering
  \begin{tabular}{ccccccc}
    \toprule
    &\multicolumn{3}{c}{Easy}    &  \multicolumn{3}{c}{Hard}    \\
    \cmidrule(r){2-4}              \cmidrule(r){5-7}
    method     & SR$\uparrow$  ($\%$)   & nodes$\downarrow$   & time(s)  %
    & SR$\uparrow$  ($\%$)   & nodes$\downarrow$  & time(s)  \\ %
    \midrule
     RRT           & $91.2$ & $37.4$ & $\mathbf{42.3}$ %
                    & $68.1$ & $81.2$ & $\mathbf{43.3}$ \\ %
    s\ourmethodabbr    & $\mathbf{97.7}$ & $\mathbf{24.1}$ & $45.8$ %
                    & $\mathbf{84.6}$ & $\mathbf{54.6}$ & $45.3$ \\ %
     sIL-steer     & $89.4$ & $43.0$ & $55.2$ %
                    & $58.2$ & $99.6$ & $80.1$ \\ %
     sRL-steer     & $90.3$ & $40.4$ & $51.5$ %
                    & $60.8$ & $94.0$ & $73.4$ \\ %
     sOpt-steer    & $84.6$ & $51.2$ & $65.7$ %
                    & $64.5$ & $93.2$ & $70.6$ \\ %
    \bottomrule
  \end{tabular}
\end{subtable}
\\
    \vspace{1.em}
    \begin{subtable}{0.5\textwidth}
  \caption{Results on 7-DoF Franka Panda robot.}
  \vspace{-1pt}
  \label{tab:state-panda}
  \centering
  \begin{tabular}{ccccccc}
    \toprule
    &\multicolumn{3}{c}{Easy}    &  \multicolumn{3}{c}{Hard}    \\
    \cmidrule(r){2-4}              \cmidrule(r){5-7}
    method     & SR$\uparrow$  ($\%$)   & nodes$\downarrow$   & time(s)  %
    & SR$\uparrow$  ($\%$)   & nodes$\downarrow$  & time(s)  \\ %
    \midrule
     RRT           & $85.7$ & $112.3$ & $166.0$ %
                    & $62.8$ & $252.5$ & $\mathbf{278.6}$ \\ %
    s\ourmethodabbr    & $\mathbf{92.0}$ & $\mathbf{67.5}$ & $\mathbf{160.5}$ %
                    & $\mathbf{76.1}$ & $162.5$ & $345.8$ \\ %
    hCBF-steer     & $39.2$ & $134.1$ & $472.6$
            & $26.3$ & $\mathbf{160.5}$ & $525.3$ \\
    sIL-steer     & $39.1$ & $324.9$ & $459.0$ %
                    & $25.3$ & $400.5$ & $547.2$ \\ %
    sRL-steer    & $83.9$ & $124.9$ & $235.9$ %
                    & $60.1$ & $266.7$ & $395.3$ \\ %
    sOpt-steer    & $26.5$ & $155.3$ & $902.8$ %
                    & $16.4$ & $174.9$ & $942.3$ \\ %
    \bottomrule
  \end{tabular}
\end{subtable}

    }
    \vspace{-2pt}
\end{table}

\noindent\textbf{Performance of LiDAR-based methods.} 
\begin{figure}[bt]
\vspace{-4pt}
    \centering
    \begin{subfigure}[t]{0.45\textwidth}
        \centering
        \includegraphics[width=\textwidth]{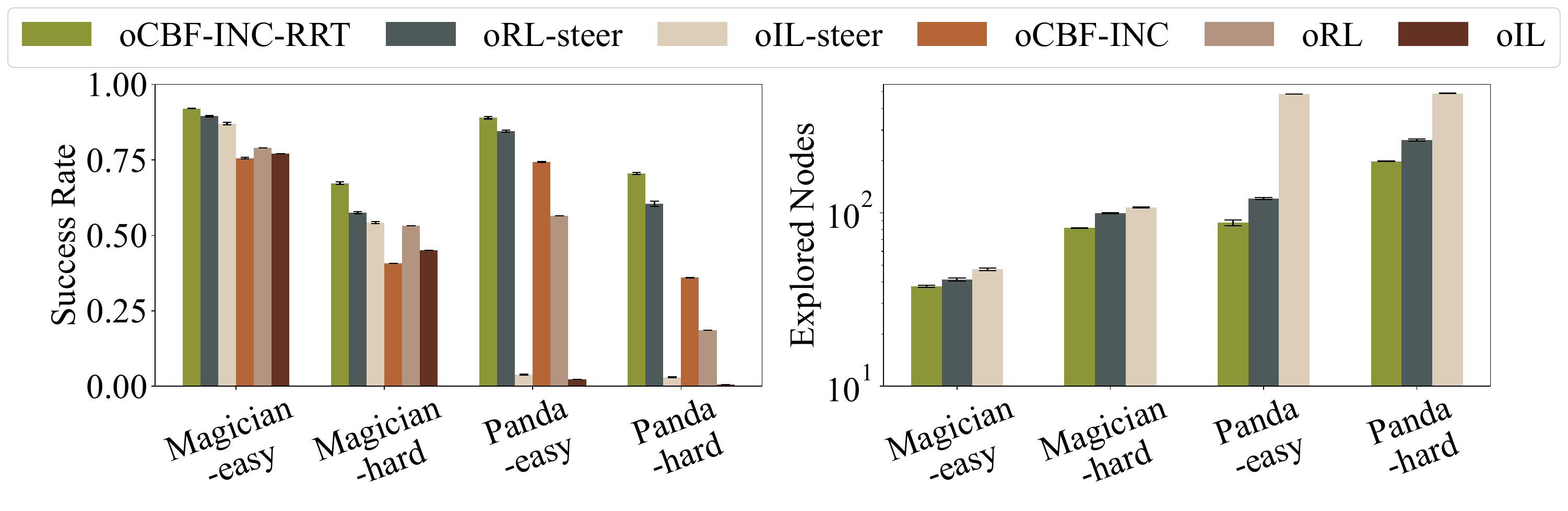}
    \end{subfigure}%
    \vspace{-2pt}
    \caption{Motion planning experiments under LiDAR-based setting. 
    }
    \label{fig:obs_whole}
    \vspace{-14pt}
\end{figure}
We evaluate algorithms that integrate various controllers into steer functions (e.g., {o\ourmethodabbr}) and those that solely leverage controllers to address the planning problems (e.g., {o\ourcontrollerabbr}). The experiments are conducted in fully-observable environments. In Fig.~\ref{fig:obs_whole}, we show that motion planning methods significantly outperform end-to-end controllers regarding success rate, demonstrating that motion planning can help improve the feasibility of finding a solution under QP formulations. Among all the motion planning methods, o\ourmethodabbr\ outperforms oRL-steer and oIL-steer on success rate and number of explored nodes, especially in hard tests. Regarding the total time consumption, our method does take a slightly longer time than vanilla RRT due to frequent inference calls of neural network. However, our method takes about $0.15\,\text{s}$ and $0.32\,\text{s}$ on average to compute the control signals and step the simulation for a 1-second period on Dobot Magician and Franka Panda, respectively. This indicates our planning can be performed faster than real-time, further establishing applicability to the real world.

\vspace{-2pt}
\subsection{Ablation Study in Simulation}

\begin{figure}[!t]
    \centering
    \begin{subfigure}[t]{0.24\textwidth}
        \centering
        \includegraphics[width=\textwidth]{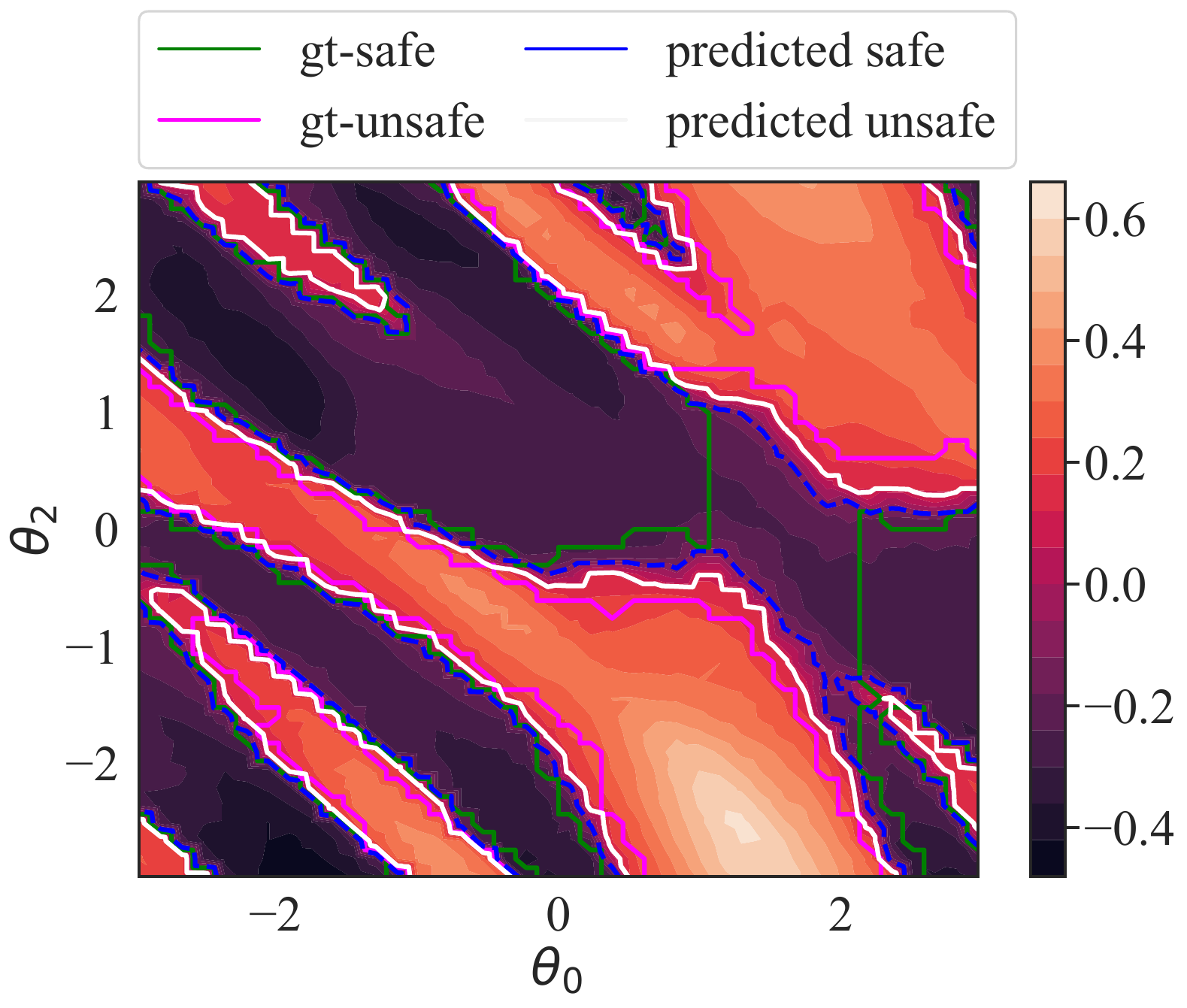}
    \end{subfigure}%
    ~
    \begin{subfigure}[t]{0.19\textwidth}
        \centering
        \includegraphics[width=\textwidth]{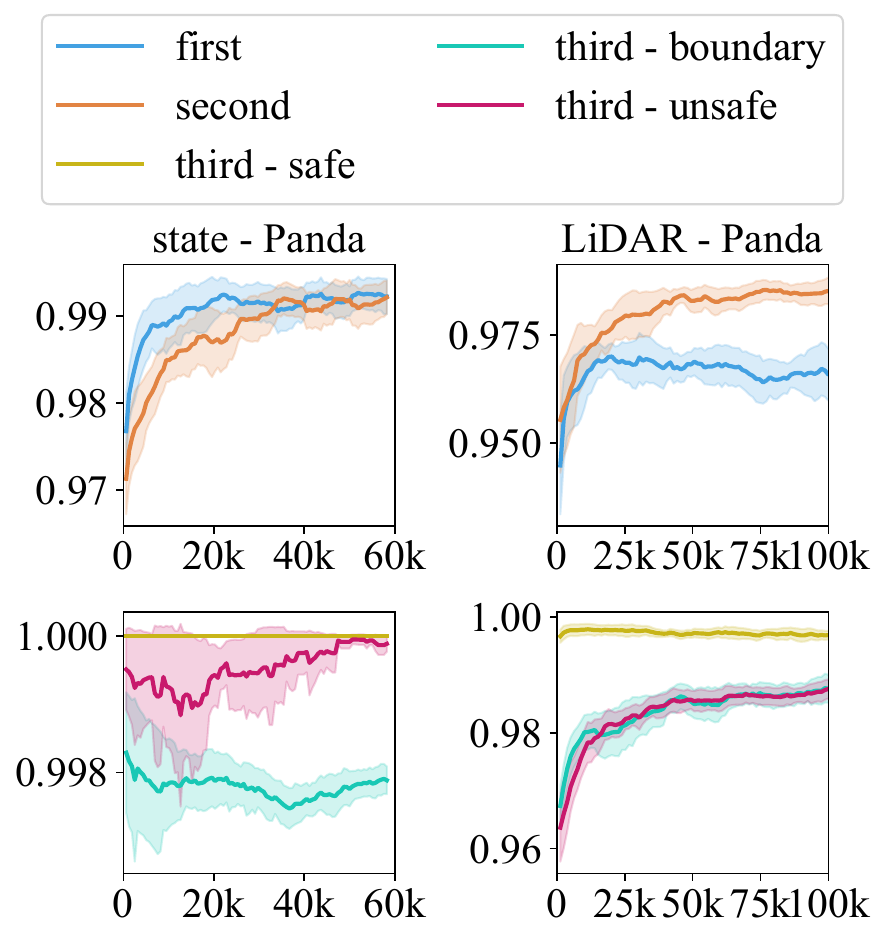}
    \end{subfigure}%
    \caption{Left: Slices of o\ournnabbr\ value for Panda, obtained by sweeping across two joints, with all other joint states and obstacle positions held constant. Right: Learning curves of constraint satisfaction rate on Panda.}
    \label{fig:learned-cbf}
    \vspace{-12pt}
\end{figure}

We first visualize CBF contour in configuration space in Fig~\ref{fig:learned-cbf}. We also plot the learning curves of satisfaction rates of each CBF constraint on our \ourmethod. All the constraints are satisfied over $99\%$ and $97\%$ on the validation sets for state-based and LiDAR-based settings after training, respectively.

We then evaluate our controller o\ourcontrollerabbr, by unrolling its control output without integrating it into the motion planning framework, over 1000 planning problems with $6$ obstacles. This experiment showcases how different neural controllers balance goal-reaching and safety.
We conduct experiments in two distinct environments: 
\textbf{(\romannum{1})} the environment is static and fully observable, where the observation contains $1024$ points uniformly sampled on the surfaces of obstacles;
\textbf{(\romannum{2})} obstacles are dynamic and move at a constant speed 
and the environment is only partially observable, where the observation is acquired by two 3D LiDARs mounted on the manipulators.

\noindent\textbf{Evaluation metrics.} We evaluate the end-to-end controllers based on three measures averaged over the testing problems: 
(1) goal-reaching rate: A goal configuration is identified as reached, only if the agent does not encounter any collision during the rollout, and eventually reaches the goal within a limited time horizon. 
(2) safety rate: the ratio of collision-free states along the entire trajectory. 
(3) makespan: rollout steps for succeeded cases.

\begin{figure}[!t]
    \centering
    \begin{subfigure}[t]{0.46\textwidth}
        \centering
        \includegraphics[width=\textwidth]{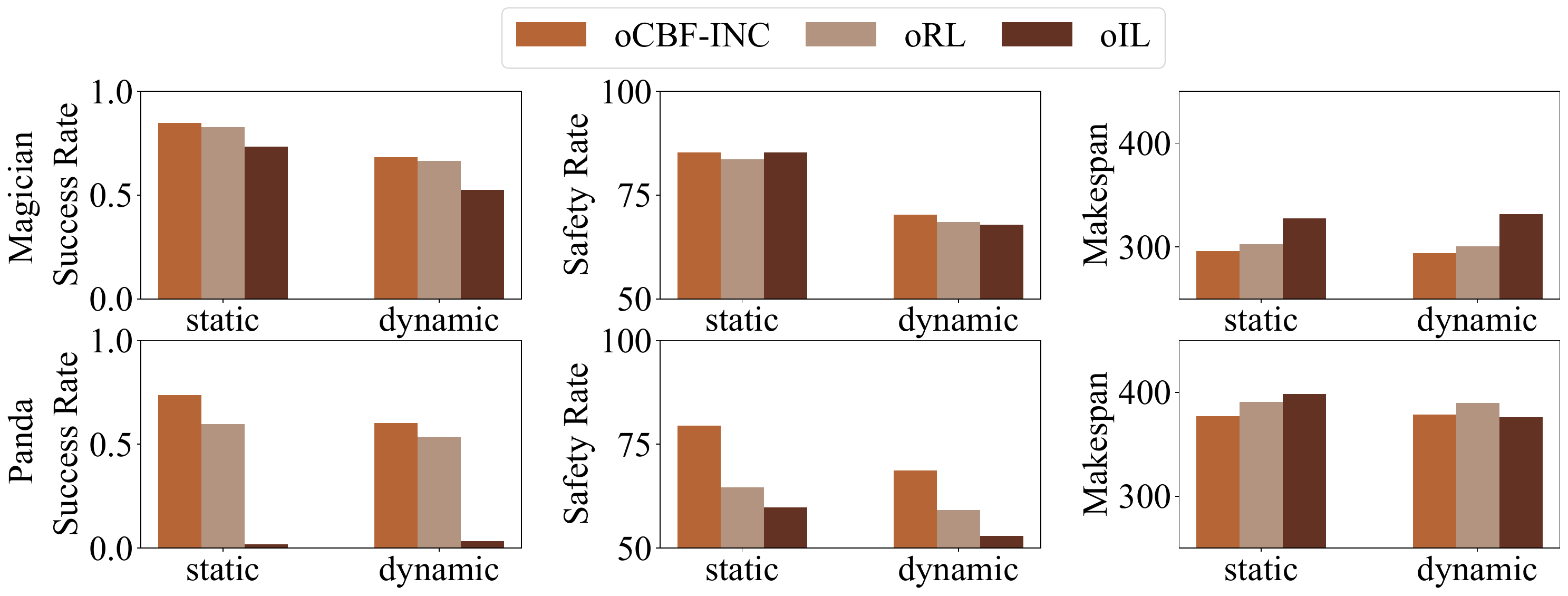}
    \end{subfigure}%
    \caption{Safety and goal-reaching performance of LiDAR-based controllers in an end-to-end manner (without motion planning module).}
    \label{fig:obs-e2e}
    \vspace{-6pt}
    \vspace{-4pt}
\end{figure}

\noindent\textbf{Performance.}
Fig~\ref{fig:obs-e2e} shows the performance of both Dobot Magician and Franka Panda robots in the considered environments. 
In the static and fully observable environment, o\ourcontrollerabbr\ outperforms baselines by approximately $2\%$ on the goal-reaching rate and safety rate for the Magician robot and by more than $15\%$ for Franka Panda. 
Although all algorithms face a substantial performance drop in the dynamic and partial-observable environment, our controller still notably exceeds oRL and oIL baselines. 
The makespan performances are generally comparable across algorithms, while o\ourcontrollerabbr\ is slightly better. This demonstrates the method achieves a great balance between efficiency and safety.

\vspace{-1pt}
\subsection{Hardware Demonstration}
Finally, we validate our proposed method on a real Franka Emika Panda controlled at $30\,\text{Hz}$, the same as in the simulation. We randomly select several planning problems in~\ref{sec:planning_result}. In order to avoid the floating blocks, we construct the obstacles and vision modules in the simulation, then synthesize real-world video with simulated obstacles, similar to~\cite{dai2023safe}. The components are communicated via ROS.

\begin{figure}[!h]
    \centering
    \begin{subfigure}[t]{0.16\textwidth}
        \centering
        \includegraphics[width=\textwidth]{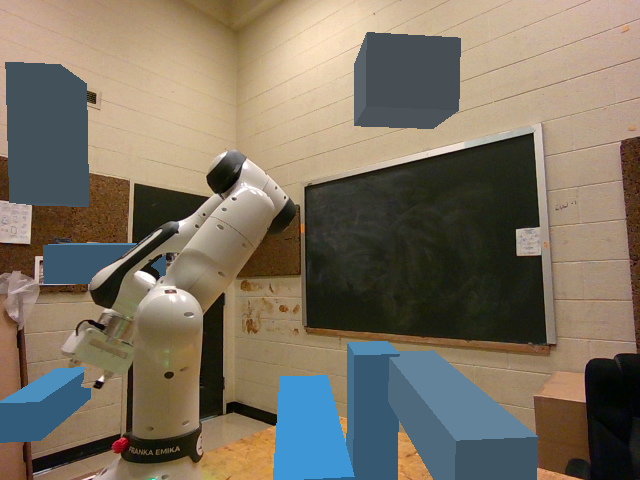}
    \end{subfigure}%
    ~ 
    \begin{subfigure}[t]{0.16\textwidth}
        \centering
        \includegraphics[width=\textwidth]{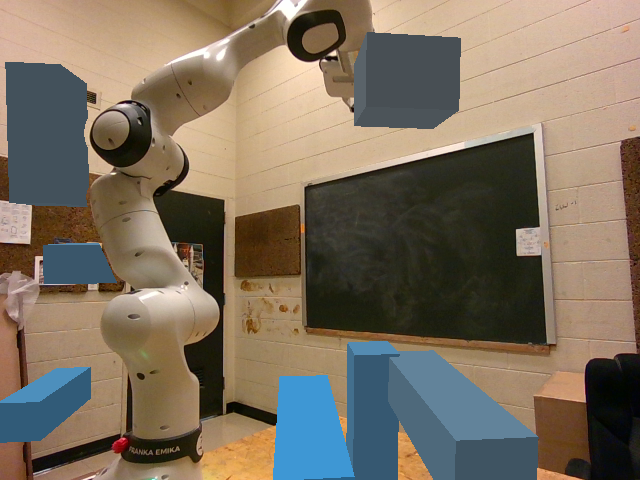}
    \end{subfigure}%
    ~ 
    \begin{subfigure}[t]{0.14\textwidth}
        \centering
        \includegraphics[width=\textwidth]{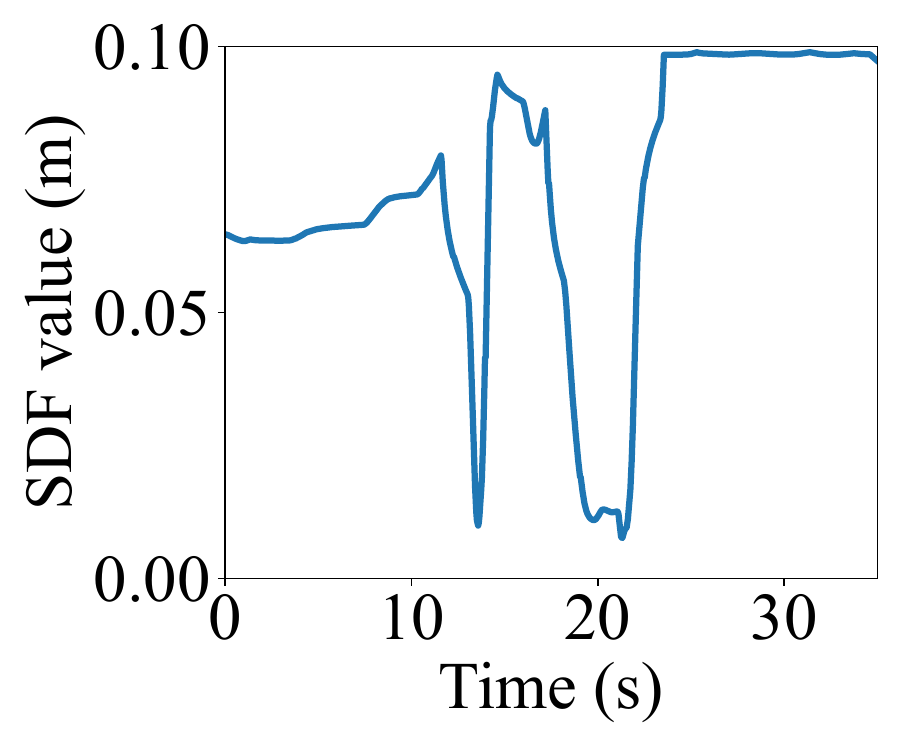}
    \end{subfigure}%
    \caption{Left \& middle: snapshots of solving a motion planning problem with our method. Right: the curve of 
    minimum signed distance to all obstacles along a trajectory. Videos are included in the supplementary.}
    \label{fig:hardware}
    \vspace{-8pt}
\end{figure}

As shown in Fig~\ref{fig:hardware} and supplementary video, our method solves the planning problems successfully. On the right of Fig~\ref{fig:hardware}, we also show the signed distance of the robot to the environment. The experiments confirm that the planned trajectory is safe and robust to the noise in execution.

\section{Discussions and Conclusion}
    This paper explores a direction for robotic safety control by integrating \ourcontroller\ - \ourcontrollerabbr\ into motion planning. 
Instead of looking for a certified CBF, we train \ournnabbr\ for robotic manipulators under different observation settings and incorporate the synthesized controller into sampling-based motion planning algorithms. 
We evaluate the proposed methods in various environments, including 4-DoF and 7-DoF arms, and in the real world. We demonstrate that \ourcontrollerabbr\ generalizes well to unseen scenarios and the overall framework outperforms other methods in terms of goal-reaching rate and exploration efficiency. 

However, there are several limitations of our paper:
(1) Since o\ournnabbr\ takes the raw sensor data as input, the performance is directly dependent on the sensor data quality. Quantifying the input cloud's uncertainty precisely remains an open-ended problem. 
(2) o\ournnabbr\ requires transforming the input point cloud into each link frame, which poses potential scalability issues for robots with higher DoF.
(3) Our computation of the Lie derivative assumes the moving speeds of obstacles are small in dynamic scenarios. We hope to relax this assumption in future work. %

\clearpage

\bibliographystyle{IEEEtran}
\bibliography{ref}  %

\begin{thebibliography}{10}
\providecommand{\url}[1]{#1}
\csname url@rmstyle\endcsname
\providecommand{\newblock}{\relax}
\providecommand{\bibinfo}[2]{#2}
\providecommand\BIBentrySTDinterwordspacing{\spaceskip=0pt\relax}
\providecommand\BIBentryALTinterwordstretchfactor{4}
\providecommand\BIBentryALTinterwordspacing{\spaceskip=\fontdimen2\font plus
\BIBentryALTinterwordstretchfactor\fontdimen3\font minus \fontdimen4\font\relax}
\providecommand\BIBforeignlanguage[2]{{%
\expandafter\ifx\csname l@#1\endcsname\relax
\typeout{** WARNING: IEEEtran.bst: No hyphenation pattern has been}%
\typeout{** loaded for the language `#1'. Using the pattern for}%
\typeout{** the default language instead.}%
\else
\language=\csname l@#1\endcsname
\fi
#2}}

\bibitem{lavalle2001rrt}
S.~M. LaValle and J.~J. Kuffner~Jr, ``Randomized kinodynamic planning,'' \emph{The international journal of robotics research}, vol.~20, no.~5, pp. 378--400, 2001.

\bibitem{kavraki1996prm}
L.~Kavraki, P.~Svestka, J.-C. Latombe, and M.~Overmars, ``Probabilistic roadmaps for path planning in high-dimensional configuration spaces,'' \emph{IEEE Transactions on Robotics and Automation}, vol.~12, no.~4, pp. 566--580, 1996.

\bibitem{webb2013kinodynamic}
D.~J. Webb and J.~Van Den~Berg, ``Kinodynamic rrt*: Asymptotically optimal motion planning for robots with linear dynamics,'' in \emph{2013 IEEE international conference on robotics and automation}.\hskip 1em plus 0.5em minus 0.4em\relax IEEE, 2013, pp. 5054--5061.

\bibitem{kala2013rapidly}
R.~Kala, ``Rapidly exploring random graphs: motion planning of multiple mobile robots,'' \emph{Advanced Robotics}, vol.~27, no.~14, pp. 1113--1122, 2013.

\bibitem{brunner2013hierarchical}
M.~Brunner, B.~Br{\"u}ggemann, and D.~Schulz, ``Hierarchical rough terrain motion planning using an optimal sampling-based method,'' in \emph{2013 IEEE International Conference on Robotics and Automation}.\hskip 1em plus 0.5em minus 0.4em\relax IEEE, 2013, pp. 5539--5544.

\bibitem{wang2020neural}
J.~Wang, W.~Chi, C.~Li, C.~Wang, and M.~Q.-H. Meng, ``Neural rrt*: Learning-based optimal path planning,'' \emph{IEEE Transactions on Automation Science and Engineering}, vol.~17, no.~4, pp. 1748--1758, 2020.

\bibitem{ichter2020learned}
B.~Ichter, E.~Schmerling, T.-W.~E. Lee, and A.~Faust, ``Learned critical probabilistic roadmaps for robotic motion planning,'' in \emph{2020 IEEE International Conference on Robotics and Automation (ICRA)}.\hskip 1em plus 0.5em minus 0.4em\relax IEEE, 2020, pp. 9535--9541.

\bibitem{AaronAmes2014}
A.~D. Ames, J.~W. Grizzle, and P.~Tabuada, ``Control barrier function based quadratic programs with application to adaptive cruise control,'' in \emph{53rd IEEE Conference on Decision and Control}, 2014, pp. 6271--6278.

\bibitem{ames2016control}
A.~D. Ames, X.~Xu, J.~W. Grizzle, and P.~Tabuada, ``Control barrier function based quadratic programs for safety critical systems,'' \emph{IEEE Transactions on Automatic Control}, vol.~62, no.~8, pp. 3861--3876, 2016.

\bibitem{manjunath2021safe}
A.~Manjunath and Q.~Nguyen, ``Safe and robust motion planning for dynamic robotics via control barrier functions,'' in \emph{2021 60th IEEE Conference on Decision and Control (CDC)}.\hskip 1em plus 0.5em minus 0.4em\relax IEEE, 2021, pp. 2122--2128.

\bibitem{yang2019cbfrrt}
G.~Yang, B.~Vang, Z.~Serlin, C.~Belta, and R.~Tron, ``Sampling-based motion planning via control barrier functions,'' in \emph{Proceedings of the 2019 3rd International Conference on Automation, Control and Robots}, 2019, pp. 22--29.

\bibitem{ahmad2022adaptive}
A.~Ahmad, C.~Belta, and R.~Tron, ``Adaptive sampling-based motion planning with control barrier functions,'' in \emph{2022 IEEE 61st Conference on Decision and Control (CDC)}.\hskip 1em plus 0.5em minus 0.4em\relax IEEE, 2022, pp. 4513--4518.

\bibitem{yang2023efficient}
G.~Yang, M.~Cai, A.~Ahmad, C.~Belta, and R.~Tron, ``Efficient lqr-cbf-rrt*: Safe and optimal motion planning,'' \emph{arXiv preprint arXiv:2304.00790}, 2023.

\bibitem{wu2016safety}
G.~Wu and K.~Sreenath, ``Safety-critical control of a planar quadrotor,'' in \emph{2016 American control conference (ACC)}.\hskip 1em plus 0.5em minus 0.4em\relax IEEE, 2016, pp. 2252--2258.

\bibitem{nguyen20163d}
Q.~Nguyen, A.~Hereid, J.~W. Grizzle, A.~D. Ames, and K.~Sreenath, ``3d dynamic walking on stepping stones with control barrier functions,'' in \emph{2016 IEEE 55th Conference on Decision and Control (CDC)}.\hskip 1em plus 0.5em minus 0.4em\relax IEEE, 2016, pp. 827--834.

\bibitem{singletary2022food}
A.~Singletary, W.~Guffey, T.~G. Molnar, R.~Sinnet, and A.~D. Ames, ``Safety-critical manipulation for collision-free food preparation,'' \emph{IEEE Robotics and Automation Letters}, vol.~7, no.~4, pp. 10\,954--10\,961, 2022.

\bibitem{dai2023safe}
B.~Dai, R.~Khorrambakht, P.~Krishnamurthy, V.~Gon{\c{c}}alves, A.~Tzes, and F.~Khorrami, ``Safe navigation and obstacle avoidance using differentiable optimization based control barrier functions,'' \emph{arXiv preprint arXiv:2304.08586}, 2023.

\bibitem{landi2019ecc}
C.~T. Landi, F.~Ferraguti, S.~Costi, M.~Bonfè, and C.~Secchi, ``Safety barrier functions for human-robot interaction with industrial manipulators,'' in \emph{2019 18th European Control Conference (ECC)}, 2019, pp. 2565--2570.

\bibitem{cheol1994two}
C.~Chang, M.~J. Chung, and B.~H. Lee, ``Collision avoidance of two general robot manipulators by minimum delay time,'' \emph{IEEE Transactions on Systems, Man, and Cybernetics}, vol.~24, no.~3, pp. 517--522, 1994.

\bibitem{rimon1997obstacle}
E.~Rimon and S.~P. Boyd, ``Obstacle collision detection using best ellipsoid fit,'' \emph{Journal of Intelligent and Robotic Systems}, vol.~18, pp. 105--126, 1997.

\bibitem{lin20166dof}
H.-C. Lin, Y.~Fan, T.~Tang, and M.~Tomizuka, ``Human guidance programming on a 6-dof robot with collision avoidance,'' in \emph{2016 IEEE/RSJ International Conference on Intelligent Robots and Systems (IROS)}, 2016, pp. 2676--2681.

\bibitem{singletary2019manipulator}
A.~Singletary, P.~Nilsson, T.~Gurriet, and A.~D. Ames, ``Online active safety for robotic manipulators,'' in \emph{2019 IEEE/RSJ International Conference on Intelligent Robots and Systems (IROS)}, 2019, pp. 173--178.

\bibitem{singletary2022kinematic}
A.~Singletary, S.~Kolathaya, and A.~D. Ames, ``Safety-critical kinematic control of robotic systems,'' \emph{IEEE Control Systems Letters}, vol.~6, pp. 139--144, 2022.

\bibitem{castaneda2021gaussian}
F.~Castaneda, J.~J. Choi, B.~Zhang, C.~J. Tomlin, and K.~Sreenath, ``Gaussian process-based min-norm stabilizing controller for control-affine systems with uncertain input effects and dynamics,'' in \emph{2021 American Control Conference (ACC)}.\hskip 1em plus 0.5em minus 0.4em\relax IEEE, 2021, pp. 3683--3690.

\bibitem{sun2021learning}
D.~Sun, S.~Jha, and C.~Fan, ``Learning certified control using contraction metric,'' in \emph{Conference on Robot Learning}.\hskip 1em plus 0.5em minus 0.4em\relax PMLR, 2021, pp. 1519--1539.

\bibitem{qin2021learning}
Z.~Qin, K.~Zhang, Y.~Chen, J.~Chen, and C.~Fan, ``Learning safe multi-agent control with decentralized neural barrier certificates,'' \emph{arXiv preprint arXiv:2101.05436}, 2021.

\bibitem{dawson2022hybrid}
C.~Dawson, B.~Lowenkamp, D.~Goff, and C.~Fan, ``Learning safe, generalizable perception-based hybrid control with certificates,'' \emph{IEEE Robotics and Automation Letters}, vol.~7, no.~2, pp. 1904--1911, 2022.

\bibitem{dean2020robust}
S.~Dean, N.~Matni, B.~Recht, and V.~Ye, ``Robust guarantees for perception-based control,'' in \emph{Learning for Dynamics and Control}.\hskip 1em plus 0.5em minus 0.4em\relax PMLR, 2020, pp. 350--360.

\bibitem{Khatib1985realtime}
O.~Khatib, ``Real-time obstacle avoidance for manipulators and mobile robots,'' in \emph{Proceedings. 1985 IEEE International Conference on Robotics and Automation}, vol.~2, 1985, pp. 500--505.

\bibitem{Santis2007self}
A.~De~Santis, A.~Albu-Schaffer, C.~Ott, B.~Siciliano, and G.~Hirzinger, ``The skeleton algorithm for self-collision avoidance of a humanoid manipulator,'' in \emph{2007 IEEE/ASME international conference on advanced intelligent mechatronics}, 2007, pp. 1--6.

\bibitem{flacco2012depth}
F.~Flacco, T.~Kröger, A.~De~Luca, and O.~Khatib, ``A depth space approach to human-robot collision avoidance,'' in \emph{2012 IEEE International Conference on Robotics and Automation}, 2012, pp. 338--345.

\bibitem{holmes2020reachable}
P.~Holmes, S.~Kousik, B.~Zhang, D.~Raz, C.~Barbalata, M.~Johnson-Roberson, and R.~Vasudevan, ``Reachable sets for safe, real-time manipulator trajectory design,'' \emph{arXiv preprint arXiv:2002.01591}, 2020.

\bibitem{wieland2007constructive}
P.~Wieland and F.~Allg{\"o}wer, ``Constructive safety using control barrier functions,'' \emph{IFAC Proceedings Volumes}, vol.~40, no.~12, pp. 462--467, 2007.

\bibitem{haviland2020neo}
J.~{Haviland} and P.~{Corke}, ``Neo: A novel expeditious optimisation algorithm for reactive motion control of manipulators,'' \emph{IEEE Robotics and Automation Letters}, vol.~6, no.~2, pp. 1043--1050, 2021.

\bibitem{pham2018optlayer}
T.-H. Pham, G.~De~Magistris, and R.~Tachibana, ``Optlayer-practical constrained optimization for deep reinforcement learning in the real world,'' in \emph{2018 IEEE International Conference on Robotics and Automation (ICRA)}.\hskip 1em plus 0.5em minus 0.4em\relax IEEE, 2018, pp. 6236--6243.

\bibitem{nadia2023sdf}
M.~Koptev, N.~Figueroa, and A.~Billard, ``Neural joint space implicit signed distance functions for reactive robot manipulator control,'' \emph{IEEE Robotics and Automation Letters}, vol.~8, no.~2, pp. 480--487, 2023.

\bibitem{liu2022safe}
P.~Liu, K.~Zhang, D.~Tateo, S.~Jauhri, Z.~Hu, J.~Peters, and G.~Chalvatzaki, ``Safe reinforcement learning of dynamic high-dimensional robotic tasks: navigation, manipulation, interaction,'' \emph{arXiv preprint arXiv:2209.13308}, 2022.

\bibitem{mcilvanna2022reinforcement}
S.~McIlvanna, N.~N. Minh, Y.~Sun, M.~Van, and W.~Naeem, ``Reinforcement learning-enhanced control barrier functions for robot manipulators,'' \emph{arXiv preprint arXiv:2211.11391}, 2022.

\bibitem{ichter2018learning}
B.~Ichter, J.~Harrison, and M.~Pavone, ``Learning sampling distributions for robot motion planning,'' in \emph{2018 IEEE International Conference on Robotics and Automation (ICRA)}.\hskip 1em plus 0.5em minus 0.4em\relax IEEE, 2018, pp. 7087--7094.

\bibitem{jurgenson2019harnessing}
T.~Jurgenson and A.~Tamar, ``Harnessing reinforcement learning for neural motion planning,'' \emph{arXiv preprint arXiv:1906.00214}, 2019.

\bibitem{strudel2021obstaclerep}
R.~Strudel, R.~G. Pinel, J.~Carpentier, J.-P. Laumond, I.~Laptev, and C.~Schmid, ``Learning obstacle representations for neural motion planning,'' in \emph{Conference on Robot Learning}.\hskip 1em plus 0.5em minus 0.4em\relax PMLR, 2021, pp. 355--364.

\bibitem{zhang2022learning}
R.~Zhang, C.~Yu, J.~Chen, C.~Fan, and S.~Gao, ``Learning-based motion planning in dynamic environments using gnns and temporal encoding,'' in \emph{Advances in Neural Information Processing Systems}, 2022.

\bibitem{janson2015fast}
L.~Janson, E.~Schmerling, A.~Clark, and M.~Pavone, ``Fast marching tree: A fast marching sampling-based method for optimal motion planning in many dimensions,'' \emph{The International journal of robotics research}, vol.~34, no.~7, pp. 883--921, 2015.

\bibitem{persson2014sampling}
S.~M. Persson and I.~Sharf, ``Sampling-based a* algorithm for robot path-planning,'' \emph{The International Journal of Robotics Research}, vol.~33, no.~13, pp. 1683--1708, 2014.

\bibitem{yu2021reducing}
C.~Yu and S.~Gao, ``Reducing collision checking for sampling-based motion planning using graph neural networks,'' \emph{Advances in Neural Information Processing Systems}, vol.~34, pp. 4274--4289, 2021.

\bibitem{zhang2018learning}
C.~Zhang, J.~Huh, and D.~D. Lee, ``Learning implicit sampling distributions for motion planning,'' in \emph{2018 IEEE/RSJ International Conference on Intelligent Robots and Systems (IROS)}.\hskip 1em plus 0.5em minus 0.4em\relax IEEE, 2018, pp. 3654--3661.

\bibitem{pfeiffer2017perception}
M.~Pfeiffer, M.~Schaeuble, J.~Nieto, R.~Siegwart, and C.~Cadena, ``From perception to decision: A data-driven approach to end-to-end motion planning for autonomous ground robots,'' in \emph{2017 ieee international conference on robotics and automation (icra)}.\hskip 1em plus 0.5em minus 0.4em\relax IEEE, 2017, pp. 1527--1533.

\bibitem{qureshi2019motion}
A.~H. Qureshi, A.~Simeonov, M.~J. Bency, and M.~C. Yip, ``Motion planning networks,'' in \emph{2019 International Conference on Robotics and Automation (ICRA)}.\hskip 1em plus 0.5em minus 0.4em\relax IEEE, 2019, pp. 2118--2124.

\bibitem{wang2022ensuring}
X.~Wang, ``Ensuring safety of learning-based motion planners using control barrier functions,'' \emph{IEEE Robotics and Automation Letters}, vol.~7, no.~2, pp. 4773--4780, 2022.

\bibitem{huang2020survey}
X.~Huang, D.~Kroening, W.~Ruan, J.~Sharp, Y.~Sun, E.~Thamo, M.~Wu, and X.~Yi, ``A survey of safety and trustworthiness of deep neural networks: Verification, testing, adversarial attack and defence, and interpretability,'' \emph{Computer Science Review}, vol.~37, p. 100270, 2020.

\bibitem{qi2017pointnet}
C.~R. Qi, H.~Su, K.~Mo, and L.~J. Guibas, ``Pointnet: Deep learning on point sets for 3d classification and segmentation,'' in \emph{Proceedings of the IEEE conference on computer vision and pattern recognition}, 2017, pp. 652--660.

\bibitem{lillicrap2015continuous}
T.~P. Lillicrap, J.~J. Hunt, A.~Pritzel, N.~Heess, T.~Erez, Y.~Tassa, D.~Silver, and D.~Wierstra, ``Continuous control with deep reinforcement learning,'' \emph{arXiv preprint arXiv:1509.02971}, 2015.

\bibitem{torabi2018behavioral}
F.~Torabi, G.~Warnell, and P.~Stone, ``Behavioral cloning from observation,'' \emph{arXiv preprint arXiv:1805.01954}, 2018.

\bibitem{gammell2014bit}
J.~D. Gammell, S.~S. Srinivasa, and T.~D. Barfoot, ``Bit*: Batch informed trees for optimal sampling-based planning via dynamic programming on implicit random geometric graphs,'' \emph{arXiv preprint arXiv:1405.5848}, 2014.

\end{thebibliography}

\end{document}